\documentclass{article}

    \PassOptionsToPackage{numbers, compress}{natbib}

    \usepackage[preprint]{neurips_2025}

\usepackage[utf8]{inputenc} 
\usepackage[T1]{fontenc}    
\usepackage{hyperref}       
\usepackage{url}            
\usepackage{booktabs}       
\usepackage{amsfonts}       
\usepackage{nicefrac}       
\usepackage{microtype}      
\usepackage{xcolor}         

\usepackage{subcaption}
\usepackage{graphicx}
\usepackage{amsmath}
\usepackage{amssymb}
\usepackage{adjustbox}
\usepackage{makecell}
\usepackage{multirow}
\usepackage{bbding}
\usepackage{color, colortbl}
\usepackage{enumitem}

\usepackage{amsthm}

\usepackage{wrapfig}

\usepackage[utf8]{inputenc} 
\usepackage[T1]{fontenc}    
\usepackage{url}            
\usepackage{booktabs}       
\usepackage{amsfonts}       
\usepackage{nicefrac}       
\usepackage{microtype}      
\usepackage{xcolor}         

\usepackage{circledsteps}

\usepackage{pifont}

\usepackage{algorithm}
\usepackage{algorithmicx}
\usepackage{algpseudocode}

\newcommand{\lin}[1]{\textcolor{black}{#1}}

\newcommand{\hk}[1]{\textcolor{black}{#1}}

\title{TokLIP: Marry Visual Tokens to CLIP\\ for Multimodal Comprehension and Generation}

\author{Haokun Lin\footnotemark[1] $^{\ 1,2,4}$, Teng Wang\footnotemark[1] $^{\ 1}$, Yixiao Ge\footnotemark[2] $^{\ 1}$, 
Yuying Ge$^{ 1}$, Zhichao Lu$^2$\vspace{0.1cm},\\
\textbf{Ying Wei$^{3}$, Qingfu Zhang$^{2}$, Zhenan Sun$^{4}$, 
Ying Shan$^{1}$}\vspace{0.3cm}\\
    $^1$ ARC Lab, Tencent PCG \quad
    $^2$ City University of Hong Kong \\
    $^3$ Zhejiang University \quad
    $^4$ NLPR \& MAIS, Institute of Automation, CAS, Beijing
    \vspace{0.15cm}\\
  \small $^*$Equal Contribution\hspace{0.2cm} $^\dagger$Corresponding Author\hspace{0.2cm} 
  \vspace{-0.5cm}
}

\begin{document}

\maketitle

\begin{abstract}
Pioneering token-based works such as Chameleon and Emu3 have established a foundation for multimodal unification but face challenges of high training computational overhead and limited comprehension performance due to a lack of high-level semantics. In this paper, we introduce TokLIP, a visual tokenizer that enhances comprehension by semanticizing vector-quantized (VQ) tokens and incorporating CLIP-level semantics while enabling end-to-end multimodal autoregressive training with standard VQ tokens. TokLIP integrates a low-level discrete VQ tokenizer with a ViT-based token encoder to capture high-level continuous semantics. Unlike previous approaches (e.g., VILA-U) that discretize high-level features, TokLIP disentangles training objectives for comprehension and generation, allowing the direct application of advanced VQ tokenizers without the need for tailored quantization operations. Our empirical results demonstrate that TokLIP achieves exceptional data efficiency, empowering visual tokens with high-level semantic understanding while enhancing low-level generative capacity, making it well-suited for autoregressive Transformers in both comprehension and generation tasks. The code and models are available at \url{https://github.com/TencentARC/TokLIP}.

\end{abstract}    
\section{Introduction}
\label{sec:intro}

\begin{figure}[!t]
    \centering
    \includegraphics[width=0.8\linewidth]{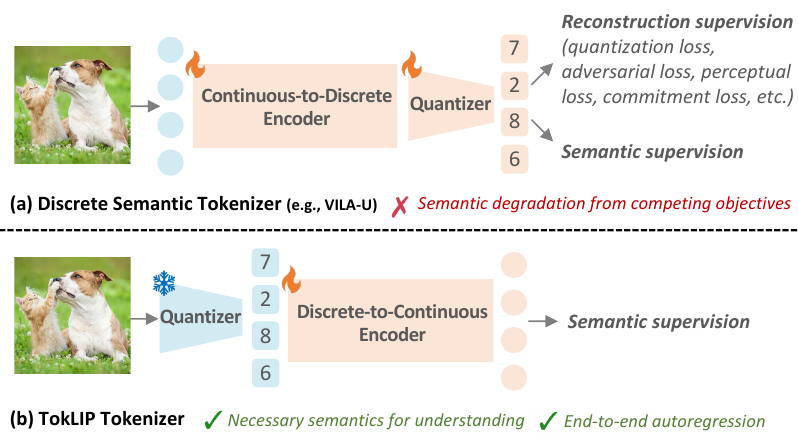}
    \caption{
\textbf{Visual tokenizers for unified multimodal comprehension and generation.}
(a) Previous discrete semantic tokenizers quantize high-level features under entangled reconstruction and semantic supervision.
(b) In contrast, our TokLIP disentangles the dual objectives by semanticizing low-level VQ tokens.
    }
    \label{fig:tokenizer_compare}
    \vspace{-0.7em}
\end{figure}

A unified autoregressive model with end-to-end next multimodal token prediction is considered the way toward the ChatGPT moment for the multimodal world. Pioneering works, such as Chameleon \citep{team2024chameleon} and Emu3 \citep{wang2024emu3}, train a single Transformer with vector-quantized (VQ) visual tokens and word tokens on multimodal sequences. Though scalable, the models suffer from expensive computational overhead for training to converge and weak comprehension performance.

VILA-U \textit{et al.} \citep{wu2024vila-u,qu2024tokenflow,zhao2025qlip} highlight that the issue lies in the fact that VQ tokens lack high-level semantics, which CLIP \citep{radford2021clip} possesses and has proven effective for LLaVA \citep{liu2023llava,liu2024llava1.5}-like frameworks designed solely for comprehension purposes.
They attempt to solve the problem via discretizing high-level semantical features, that is, training a continuous-to-discrete visual tokenizer with both reconstruction and text alignment objectives, illustrated in Figure~\ref{fig:tokenizer_compare} (a).
However, the conflict between two distinct objectives and the information loss of visual semantics caused by quantization operation pose new challenges for properly unifying multimodal comprehension and generation.

The evidence suggests that a) the objectives of multimodal comprehension and generation should be disentangled due to varying semantic demands, and b) improved quantization techniques yield superior results. 
To this end, we argue that the proper approach to meet such principles is to \textit{semanticize VQ tokens} rather than discretize CLIP features.
In this paper, we introduce \textbf{TokLIP}, a discrete-to-continuous visual tokenizer as shown in Figure~\ref{fig:tokenizer_compare} (b) that enables end-to-end autoregressive training of a single Transformer with sequences of multimodal discrete tokens, while simultaneously incorporating CLIP-level semantics to enhance multimodal comprehension.

Specifically, the TokLIP tokenizer consists of an off-the-shelf low-level VQ tokenizer, such as VQGAN~\citep{esser2021vqgan,sun2024llamagen}, and a ViT-based token encoder with causal attention. An image is first discretized at a low level, after which the visual code embeddings are fed into the token encoder to capture high-level continuous semantics. The token encoder is pre-trained with only semantic supervision, specifically through text alignment and CLIP distillation, and is subsequently equipped with the autoregressive Transformer for end-to-end multimodal token prediction.

TokLIP offers three key advantages:  
(1) It effectively disentangles the training objectives for comprehension and generation across different semantic levels, eliminating the need for complex training strategies to balance these objectives.  
(2) It integrates CLIP-level semantics into Emu3-like unified autoregressive models, enhancing comprehension performance while preserving end-to-end multimodal autoregression on plain VQ tokens.  
(3) It leverages the strengths of VQ-based methods, enabling the direct use of state-of-the-art VQ tokenizers without addressing information loss from quantization training at the semantic level.

We empirically validate the effectiveness of TokLIP by comparing it with other tokenizers and integrating it with large language models (LLMs) for autoregressive multimodal comprehension and generation.
Our analysis reveals three key findings: 
TokLIP demonstrates exceptional image representation capabilities, utilizing less than 20\% of the pretraining data required by VILA-U, while significantly outperforming established tokenizers such as VILA-U and QLIP in zero-shot ImageNet classification. Additionally, TokLIP retains the original reconstruction capabilities of the low-level VQ tokenizer without necessitating specific optimization for generation and quantization.
(2) \textit{Multimodal comprehension at significantly reduced training cost}: When integrated with LLMs for comprehension tasks, TokLIP achieves comparable results using only 
5\% of the training data required by SynerGen-VL. 
This reveals that TokLIP unleashes the potential of discrete token-based multimodal LLMs to efficiently deliver advanced comprehension performance similar to that of the LLaVA series.
(3) \textit{Synergistic autoregressive image generation}: \lin{We demonstrate that high-level semantic features complement low-level code embeddings to enhance image generation, underscoring TokLIP's ability to integrate diverse abstraction levels for robust multi-modal outputs.}

\section{Related Works}
\label{sec:related}

\begin{figure*}[!ht]
    \centering
    \includegraphics[width=1\linewidth]{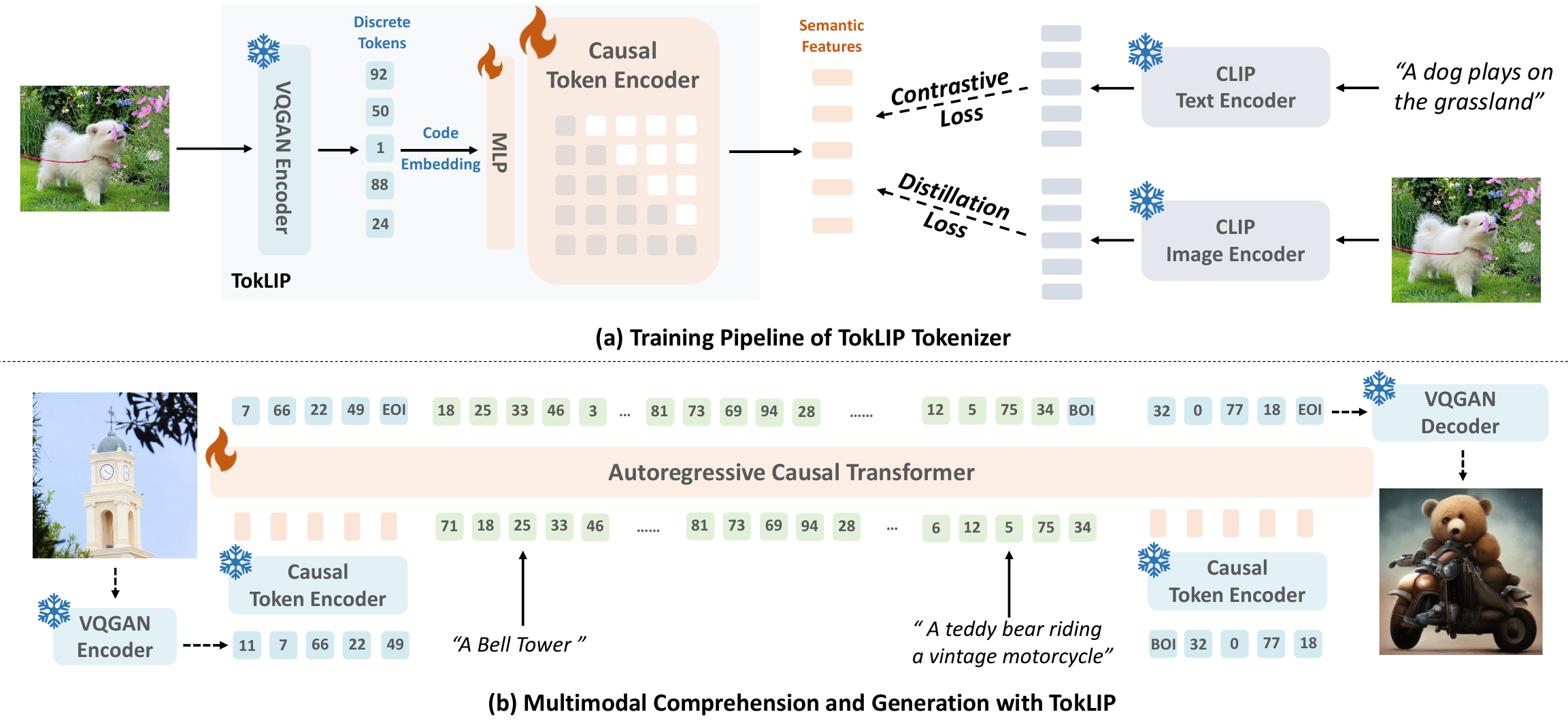}
    \caption{
\textbf{Overview of TokLIP.}
(a) Given an image, it is first quantized by an off-the-shelf VQGAN encoder before being fed into the causal token encoder equipped with a multi-layer perceptron. The token encoder is initialized from a pre-trained bidirectional CLIP vision encoder and semanticizes the discrete visual codes under a data-efficient learning strategy with contrastive loss and distillation loss.
(b) The pretrained TokLIP is further employed in the framework of LLMs to tokenize images for end-to-end multimodal autoregressive learning. Given the semantic knowledge in the pretrained token encoder, we achieve advanced vision-language comprehension results with few computational overhead. 
Furthermore, TokLIP enables autoregressive image generation on plain VQ tokens.
    }
    \label{fig:toklip-pipeline}    
\end{figure*}

\paragraph{Vector-quantized tokenizers.}
Vector quantization (VQ) is effective in processing different modalities, especially images~\citep{van2017vqvae,esser2021vqgan,yu2021vector,lee2022rqvae,chang2022maskgit,wang2025vqkd,razavi2019vqvae2,tian2025VAR}, compressing high-dimensional raw data into discrete token sequences for efficient learning.
The pioneering work VQ-VAE~\citep{van2017vqvae} proposes to discretize continuous representations by assigning them to the closest vectors in a learnable codebook. 
VQGAN~\citep{esser2021vqgan} incorporates adversarial and perceptual losses to further improve generation quality. 
Recent studies, such as LlamaGen~\citep{sun2024llamagen}, MAGVIT2~\citep{luo2024openmagvit,yu2023magvit} and VAR~\citep{tian2024var}, demonstrate the great potential of VQ tokenizers in autoregressive image generation.
Such tokenizers are naturally suited for generation tasks due to capturing low-level patterns, however, they struggle to provide necessary semantics for multimodal comprehension tasks, as evidenced in \citep{wu2024janus,xie2024show-o}. 
In this work, we present TokLIP, which semanticizes VQ tokens to enhance multimodal comprehension while maintaining end-to-end multimodal autoregressive learning on plain VQ tokens when integrated into large multimodal models.

\paragraph{Large multimodal models with unified comprehension and generation.}
Exploring unified modeling of multimodal comprehension and generation in the framework of large language models has gained significant attention.
Despite various attempts, the core challenge remains addressing the essential conflicts of comprehension and generation tasks. There are four categories of approaches to meet the challenge:
(1) \textit{External diffusion}: \citep{ge2023seed-llama,sun2023emu,tong2024metamorph,ge2024seed-x,sun2024generative,wu2023next-gpt} utilize external pretrained diffusion models linked through semantic features for image generation, resulting in independent processing of comprehension and generation.
\hk{Recently,
\citep{chen2025blip3o,pan2025metaquery,ai2025ming} explore how to extend pre-trained MLLMs with DiT blocks for generation.}
(2) \textit{Internal diffusion}: \citep{xie2024show-o,zhou2024transfusion,shi2024llamafusion,xie2025show-o2} facilitate visual-language comprehension and generation within a single model; however, their use of denoising steps during image generation tasks prevents true fusion of images and text.
(3) \textit{Dual encoders}: \citep{wu2024janus,ma2024janusflow,chen2025janus-pro} employ a continuous semantic encoder for comprehension and a discrete encoder for generation, \hk{while \citep{deng2025bagel} adopts VAE for generation and explores the scale-up of the Mixture-of-depth architecture.}
These works will incur increased computational overhead and result in independent processing of these tasks.
(4) \textit{Unified tokenizer}: \citep{liu2024lwm,team2024chameleon,wang2024emu3} pioneered this line of research by training a single autoregressive Transformer with discrete image VQ tokens and text tokens. \citep{wu2024vila-u,zhao2025qlip,han2025tar} further improve the multimodal comprehension performance by proposing new discrete visual tokenizers with high-level semantics.
\hk{\citep{qu2024tokenflow,ma2025unitok} utilize multiple codebooks to alleviate conflicts between comprehension and generation.}

It is evident that only the "unified tokenizer" category can effectively integrate comprehension and generation tasks, along with text and image tokens, into a single autoregressive model. However, state-of-the-art methods~\citep{wu2024vila-u,zhao2025qlip,ma2025unitok,qu2024tokenflow} in this area face challenges due to conflicting training objectives—namely, reconstruction losses for generation and text-alignment losses for comprehension—as well as information loss from quantization training. This complexity makes stable training particularly difficult.
We contend that it is more proper to semanticize low-level discrete visual tokens, as our proposed TokLIP, rather than to discretize high-level continuous visual features, as done in previous research~\citep{wu2024vila-u,qu2024tokenflow}.

\section{Method}
\label{sec:method}

TokLIP is a discrete-to-continuous tokenizer tailored for integrating multimodal comprehension and generation tasks into a single autoregressive model.
In this section, we outline the TokLIP architecture and training details in Section~\ref{subsec:toklip_method}, explore its application to multimodal comprehension and generation under the framework of LLMs in Section~\ref{subsec:method_und&gen}, and conclude with a discussion and comparison of its main differences with existing tokenizers in Section~\ref{subsec:discussion}.

\subsection{TokLIP Tokenizer}
\label{subsec:toklip_method}

Pioneering works~\citep{team2024chameleon,wang2024emu3} in multimodal unification adopt VQ-based low-level tokenizers to encode images for both comprehension and generation tasks. 
These fidelity-oriented tokenizers primarily capture basic image features, such as contours and patterns, making them ideal for generation tasks.
However, visual understanding tasks require high-level semantic representations to enhance interpretability and reasoning capabilities, as evidenced in \citep{liu2023llava,zhu2023minigpt,bai2023qwenvl}. 
The lack of such semantic knowledge in VQ-based tokenizers limits their effectiveness in multimodal understanding.

\paragraph{Overall architecture.}
To solve the problem, we explore the possibility of enriching VQ-based tokenizers~\citep{esser2021vqgan,sun2024llamagen} with high-level semantic knowledge. 
CLIP~\citep{radford2021clip,zhai2023siglip}, trained on hundreds of millions of image-text pairs, has demonstrated remarkable success in visual understanding. 
To this end, we propose TokLIP, a framework that integrates fidelity encoders with CLIP vision encoders to simultaneously preserve both generation and comprehension capabilities in a single tokenizer, as illustrated in Figure \ref{fig:toklip-pipeline} (a).
Specifically, given an input image $I\in \mathbb{R}^{H \times W \times 3}$, we first quantize it using a VQGAN encoder $\mathcal{E}_\text{vq}$ to obtain the code embeddings of discrete tokens $x_\text{vq}=\mathcal{E}_\text{vq}(I) \in \mathbb{R}^{ (\frac{H.W}{p.p}) \times \hat{d}}$, where $p$ is the downsample ratio and $\hat{d}$ is the dimension of VQGAN codebook embeddings. The code embeddings $x_\text{vq}$ are further mapped to CLIP input feature dimension $d$ via a multi-layer perception (\texttt{MLP}) layer before being fed into a token encoder $\mathcal{E}_\text{tok}$ which shares the same architecture as CLIP's ViT encoder but uses causal attention.
To conclude, the TokLIP tokenizer encodes the input image into semantic features $x_\text{toklip} \in  \mathbb{R}^{ (\frac{H.W}{p.p}) \times d}$ from discrete to continuous, denoted as
\begin{equation}
\label{eq:x_toklip}
    x_\text{toklip} = \mathcal{E}_\text{tok} \left( \text{MLP}(x_\text{vq}) \right) = \mathcal{E}_\text{tok} \left( \text{MLP}(\mathcal{E}_\text{vq}(I)) \right).
\end{equation}
Our objective is to train the token encoder $\mathcal{E}_\text{tok}$ along with the MLP layer on large-scale image-text data to effectively learn semantic representations from discrete token inputs.

\paragraph{Causal attention enables end-to-end multimodal autoregression.}

In the original CLIP architecture~\citep{radford2021clip}, the vision Transformer encoder employs bidirectional attention to capture the rich semantic relationships among image patches and facilitate a holistic understanding of the input. 
However, in the context of multimodal autoregressive learning, the bidirectional attention mechanism hinders the training paradigm of end-to-end next token prediction, where each token can only be predicted by attending to its preceding tokens.
To address this discrepancy, we adopt causal attention for our token encoder, which enables end-to-end next multimodal token prediction when further equipping TokLIP with LLMs as shown in Figure \ref{fig:toklip-pipeline} (b). In such a way, image and text signals, as well as comprehension and generation tasks, can be truly fused in a single causal Transformer.

\paragraph{Training objectives.}
We empirically find that directly training a causal ViT presents performance degradation compared to the original bidirectional ones, as the learning process struggles to adapt to this more constrained attention mechanism. 
We tailor a training strategy to mitigate the performance degradation.
First, 
instead of training the token encoder from scratch, we initialize it with a pretrained bidirectional CLIP vision encoder. Our ablation studies, detailed in Section~\ref{subsec:ablation}, show that inheriting knowledge from a well-trained CLIP model significantly reduces the performance gap between causal and bidirectional models.
Next, we combine contrastive loss $\mathcal{L}_\text{contra}$ with distillation loss $\mathcal{L}_\text{distill}$ as the training objectives. 
Specifically, given an image-text pair $(x_i,y_i)$ in a mini-batch $(\mathbb{X}, \mathbb{Y})$, we obtain the image and text representations $z^{i}_\text{img}, z^i_\text{text}$ from pretrained CLIP teacher encoders. We denote the \texttt{[CLS]} token of $x^i_\text{toklip}$ (Eq. (\ref{eq:x_toklip})) as $z^i_\text{toklip}$.
The contrastive loss and distillation loss are formulated as follows:
\begin{align}
    \mathcal{L}_\text{contra}(x_i, y_i) &= \text{InfoNCE} \left(z^{i}_\text{toklip}, z^i_\text{text} \right), \nonumber\\
    \mathcal{L}_\text{distill}(x_i) &= \text{MSE} \left(z^i_\text{toklip}, z^i_\text{img} \right),
\end{align}
where the overall training objective is $\mathcal{L} = \mathcal{L}_\text{contra} + \mathcal{L}_\text{distill}$
that enables the effective and data-efficient training of TokLIP tokenizer. Only the MLP layer and the token encoder are trainable. By freezing the VQGAN encoder, we decouple the reconstruction capability from the TokLIP training process, avoiding potential objective conflicts and easing the overall optimization.

\paragraph{Extensibility.}
We emphasize that TokLIP is not a fixed model architecture but a continuously improvable framework that can be enhanced with advanced VQ-based tokenizers and powerful CLIP encoders. For instance, we can replace VQGAN with superior tokenizers such as MAGVIT-v2~\citep{yu2023magvit}, Cosmos~\citep{agarwal2025cosmos} or large-scale pre-trained tokenizers as utilized in Emu3~\citep{wang2024emu3}. Additionally, we can substitute CLIP encoders with more effective ViT-based semantic encoders like SigLIP~\citep{zhai2023siglip}, AIM~\citep{el2024aim}, or DFN~\citep{fang2023dfn}.
\lin{
Notably, TokLIP can be further extended by combining it with a diffusion head, following the success of GPT-4o~\citep{chen2025empirical-gpt4o,openai2025gpt4o}, to generate higher-quality and high-resolution images.
}

\begin{figure*}[!t]
    \centering
    \includegraphics[width=1\linewidth]{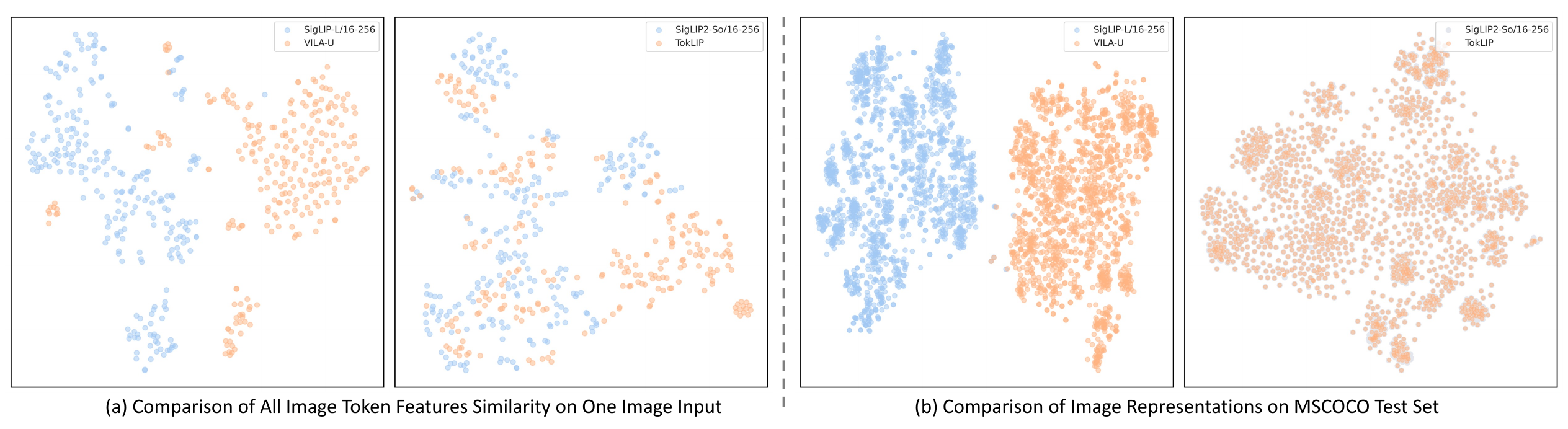}
    \caption{
\textbf{Empirical Comparison of Feature Representations.}
(a) t-SNE visualization of all tokenized features from VILA-U (left) and TokLIP (right), compared with patch features from respective SigLIP encoders for a single input image.
(b) t-SNE visualization of 1500 image representations from VILA-U (left) and TokLIP (right), compared with representations from the SigLIP encoders. The representations are achieved using an attention pooler for cross-modal retrieval tasks. 
\textit{TokLIP features remain closely aligned with SigLIP across both levels.}
    }
    \vspace{-0.7em}
    \label{fig:tsne-vis}    
\end{figure*}

\subsection{Multimodal Autoregression with TokLIP}
\label{subsec:method_und&gen}

The pretrained TokLIP tokenizer enables end-to-end multimodal autoregressive learning in a single causal Transformer, as demonstrated in Figure \ref{fig:toklip-pipeline} (b), which unleashes the great potential of multimodal comprehension and generation applications. Specifically, during the autoregressive training of the Transformer, TokLIP is frozen to tokenize images, where the output image tokens are fed into the Transformer together with the text tokens. Only the cross-entropy loss is employed here for multimodal token prediction.
In the paper, we take \textbf{image-to-text} and \textbf{text-to-image} as typical tasks for verifying the effectiveness of TokLIP in multimodal comprehension and generation applications, respectively.
\lin{We concatenate the low-level output tokens from VQGAN with the high-level output features from the causal token encoder for feature fusion and mutual enhancement.}

\subsection{Discussion}
\label{subsec:discussion}

\paragraph{Discretize CLIP or semanticize VQ?}
Another line of work aims to train a semantic tokenizer by quantizing high-level image features from the CLIP encoder into discrete tokens~\citep{wu2024vila-u,zhao2025qlip}. These methods typically rely on joint training with both reconstruction and semantic losses. 
The key distinction between these approaches and TokLIP is that the former adopts a pipeline that transitions from continuous to discrete representations (\textit{i.e.}, discretize high-level CLIP features), while TokLIP follows a discrete-to-continuous way (\textit{i.e.}, semanticize low-level VQ tokens).

Intuitively, TokLIP properly disentangles the training objectives of comprehension and generation at different semantic levels (\textit{i.e.}, low-level discrete tokens for generation and high-level continuous features for comprehension), alleviating the accuracy loss caused by objective conflicts in previous methods.
Furthermore, VQ is known to be difficult to optimize due to the complicated training objectives. In TokLIP, we can directly exploit the state-of-the-art VQ tokenizers and inherit their generation capabilities without tailoring strategies for quantization training.

To better understand the advantages of disentangled training objectives, we conduct a feature analysis for comparison, as shown in Figure \ref{fig:tsne-vis}, demonstrating that TokLIP learns better semantics than VILA-U.
We primarily compare the feature similarity between tokenizers and their initialized SigLIP~\citep{zhai2023siglip,tschannen2025siglip2} encoders. 
Specifically, we extract the normalized features $x_\text{toklip}$/$x_\text{vila-u}$ and $x_\text{siglip}$ for a single input image and compute the cosine similarity between $x_\text{toklip}$/$x_\text{vila-u}$ and $x_\text{siglip}$. 
Our results show that TokLIP achieves a similarity score of 0.83, outperforming VILA-U, which scores 0.75. 
We further visualize the normalized features using t-SNE in Figure~\ref{fig:tsne-vis} (a). 
On the left, the VILA-U features are far from the SigLIP-ViT-L/16 features, indicating that the feature spaces differ significantly. 
In contrast, TokLIP features are much closer to SigLIP2-So/16. 
These results suggest that the training process in VILA-U drastically alters the feature representations compared to the semantic encoder. 
We attribute this to the conflict between the reconstruction and semantic objectives and the information loss during the quantization process.

Additionally, we analyze the image representations used for retrieval and classification tasks, which are achieved by pooling image features $x_\text{siglip}$ through an attention pooling head in SigLIP~\citep{zhai2023siglip}. 
We randomly select 1,500 images from the MSCOCO test set and visualize these representations using t-SNE in Figure~\ref{fig:tsne-vis} (b). 
Notably, the representations from VILA-U are relatively distant from SigLIP, while TokLIP shares nearly the same distribution as SigLIP2.
This aligns with TokLIP's stronger retrieval performance demonstrated in Section~\ref{subsec:exp_tokenizer_compare}. These analyses further show that our pipeline learns more similar semantics to pre-trained CLIP ViTs, thus achieving superior comprehension performance.

\paragraph{Comparison with CLIP Encoders.}

Pretrained on large-scale image-text pairs, CLIP~\citep{radford2021clip,tschannen2025siglip2,lin2024mope,gao2024clip-adapter,ma2025genhancer} has achieved remarkable success in capturing high-level semantic features of images, making it highly effective for feature extraction in multimodal large language models (LLMs). Numerous studies~\citep{liu2023llava,zhu2023minigpt,bai2023qwenvl,lu2024deepseekvl,lin2024duquant,zhou2024doge,zhou2025scale} leverage CLIP's continuous features by combining them with text tokens in LLMs and training on vision-language data to enhance performance in comprehension tasks. However, while CLIP excels at encoding high-level semantics, it often overlooks low-level structures and textural details, which are crucial for visual generation tasks. 
Models such as SEED-X~\citep{ge2024seed-x}, DreamLLM~\citep{dong2023dreamllm}, and MetaMorph~\citep{tong2024metamorph} incorporate CLIP's continuous features with text tokens for autoregressive prediction but still rely on diffusion models~\citep{podell2023sdxl} for generation inference to compensate for the lack of low-level details. In contrast, TokLIP proposes a unified tokenizer that integrates both low-level features and high-level semantics, enabling direct tokenization with CLIP encoders and allowing for more effective handling of visual generation tasks with VQGAN decoder.
\section{Experiments}
\label{sec:exp}

\begin{table*}[!t]
    \caption{Comparison with state-of-the-art tokenizers and CLIP encoders on zero-shot ImageNet classification and MSCOCO 5K retrieval tasks.
    TR refers to image-to-text retrieval, while IR denotes text-to-image retrieval.
    }
    \vspace{-10pt}
    \begin{center}
    \renewcommand\arraystretch{1.3}
    \resizebox{0.87\linewidth}{!}{
    \begin{tabular}{llcccc}
    \toprule
    \textbf{Model}    & \textbf{Dataset} & \textbf{Res.} & \textbf{IN Top1$ \uparrow$} & \textbf{COCO TR@1$ \uparrow$} & \textbf{COCO IR@1$ \uparrow$} \\ \midrule
    \rowcolor{gray!10}
    EVA02-B/16~\citep{fang2024eva02}            & Merged-2B        & 224                 & 74.7             & 58.74              & 42.15              \\
    \rowcolor{gray!10}
    CLIP-L/14~\citep{radford2021clip}           & WIT400M          & 224                 & 75.5             & 56.34              & 36.51              \\
    \rowcolor{gray!10}
    MetaCLIP-L/14~\citep{xu2023metaclip}        & MetaCLIP400M     & 224                 & 76.2             & 60.00              & 43.81              \\
    VILA-U~\citep{wu2024vila-u}                 & COYO700M         & 256                 & 73.3             & 62.56              & 45.37              \\
    QLIP~\citep{zhao2025qlip}                   & DataComp1B       & 256                 & 74.3             & 55.62              & 38.91              \\
    \rowcolor{purple!10}
    \textbf{TokLIP-B}                             & Mix 125M         & 256                 & 76.4             & 64.06              & 48.46              \\ \midrule
    \rowcolor{gray!10}
    OpenCLIP-ViT-L/14~\citep{cherti2023openclip} & LAION2B          & 336                 & 75.3             & 63.36              & 46.51              \\
    \rowcolor{gray!10}
    SigLIP-B/16~\citep{zhai2023siglip}          & WEBIL10B         & 384                 & 76.5             & 67.74              & 49.90              \\
    \rowcolor{gray!10}
    SigLIP-So/14~\citep{zhai2023siglip}         & WEBIL10B         & 384                 & 83.2             & 70.20              & 52.00               \\
    VILA-U~\citep{wu2024vila-u}                 & COYO700M         & 384                 & 78.0             & -                  & -                  \\
    QLIP~\citep{zhao2025qlip}                   & DataComp1B       & 384                 & 79.1             & 60.86              & 43.00              \\
    \rowcolor{purple!10}
    \textbf{TokLIP-L}                             & Mix 80M          & 384                 & 80.0             & 68.00              & 52.87              \\  
    \midrule
    \rowcolor{gray!10}
    SigLIP-B/16~\citep{zhai2023siglip}          & WEBIL10B         & 512                 & 79.1             & 68.72              & 50.55              \\
    \rowcolor{gray!10}
    SigLIP2-B/16~\citep{tschannen2025siglip2}   & WEBIL10B         & 512                 & 81.2             & 71.20              & 55.20               \\ 
    \rowcolor{gray!10}
    SigLIP2-L/16~\citep{tschannen2025siglip2}   & WEBIL10B         & 512                 & 83.5             & 72.10              & 55.20               \\ 
    \rowcolor{purple!10}
    \textbf{TokLIP-XL}                          & Mix 70M          & 512                 & 80.8             & 69.40              & 53.77              \\ 
    \bottomrule
    \end{tabular}
    }
    \label{tab:understanding-comparison}
    \end{center}
    \vspace{-0.2cm}
\end{table*}

\subsection{Experimental Setup}

\paragraph{Implementation details.}

\hk{
The TokLIP architecture integrates a pre-trained text-to-image VQGAN tokenizer with a causal token encoder based on the ViT-So400M architecture. We implement three variants of TokLIP with different input resolutions: TokLIP-B (256×256), TokLIP-L (384×384), and TokLIP-XL (512×512).
TokLIP-B and TokLIP-L use the VQGAN from LlamaGen~\citep{sun2024llamagen}, while TokLIP-XL adopts IBQ~\citep{shi2024ibq} with a 260K-sized codebook. All models operate with a downsampling ratio of 16.
The token encoder is initialized from SigLIP2~\citep{tschannen2025siglip2}, providing a strong foundation for semantic understanding. A two-layer MLP bridges the VQGAN output and the token encoder, projecting the low-dimensional VQGAN features into the input space of the CLIP-based encoder.
For training, TokLIP-L and TokLIP-XL are trained on 80/70 million image-text pairs from CapsFusion~~\citep{yu2024capsfusion}, CC12M~\citep{changpinyo2021conceptual}, and LAION-high-resolution~\citep{schuhmann2022laion-5b}. TokLIP-B further includes a subset of LAION400M~\citep{schuhmann2021laion400m}, resulting in a combined dataset of 125 million samples. Low-resolution images are filtered to ensure effective model learning.
}

For multimodal comprehension, we utilize TokLIP as the input encoder and pass the output concatenated features through a 2-layer MLP to Qwen2.5-7B-Instruct \citep{yang2024qwen2.5}. 
\hk{
Our training process consists of two stages.
In Stage 1, we jointly train the MLP and the LLM on 4.5 million images with detailed captions for multimodal alignment, using data sourced from LLaVA-OneVision~\citep{li2024llava-onevison}.
In stage 2, we perform instruction tuning using the LLaVA-OneVision along with LLaVA-Next~\citep{liu2024llavanext} datasets.
For visual generation, we follow LlamaGen \citep{sun2024llamagen}, training class-to-image generation models on ImageNet. We replace LlamaGen's text-to-image VQGAN tokenizer with TokLIP.
}

\paragraph{Evaluation details.}
We evaluate the understanding capacity of our TokLIP on ImageNet classification~\citep{deng2009imagenet} and MSCOCO 5K retrieval~\citep{chen2015coco-caption} tasks.
For multimodal comprehension, we rigorously evaluate our model on several zero-shot vision-language benchmarks, which include 
POPE~\cite{li2023pope}, MME-P (MME-Perception)~\citep{fu2023mme}, MMB (MMBench)~\citep{liu2024mmbench}, SEED (SEED-Bench Img)~\citep{li2023seed-img}, GQA~\citep{hudson2019gqa}, MMMU~\citep{yue2024mmmu}, and MM-Vet~\citep{yu2023mm-vet},
covering
a diverse range of tasks that test both visual and linguistic understanding in a variety of real-world settings.
For visual generation, we mainly report Fréchet inception distance (FID) on the ImageNet 256x256 benchmark.

\begin{table*}[]
    \caption{
    Comparative analysis of tokenizers on multimodal comprehension tasks.
    We adopt the LLaVA-v1.5~\citep{liu2024llava1.5} training framework and combine tokenizers with Qwen2.5-7B-Instruct~\citep{yang2024qwen2.5}, training on LLaVA-v1.5 data.
    $^\ddagger$ denotes the results are reported from the original paper.
    }
    \begin{center}
    \resizebox{1\linewidth}{!}{
    \begin{tabular}{llcccccc}
    \toprule
        \textbf{Model} & \textbf{Data} & \textbf{Res.} & \textbf{POPE$ \uparrow$}  & \textbf{MME-P$ \uparrow$} & \textbf{SEED$ \uparrow$}  & \textbf{GQA$ \uparrow$} & \textbf{MMMU$ \uparrow$}  \\ \midrule
        Discrete Tokenizer (VQGAN)    & LLaVA-v1.5 data   &256 & 65.6 & 716.8 & 35.0 & 39.8 & 36.6 ~ \\ 
        Discrete Semantic Tokenizer (VILA-U)$^\ddagger$   & MMC4+ShareGPT4V(7M)   &256 & 83.9 & 1336.2 & 56.3  & 48.3 & -  \\
        \rowcolor{purple!10} 
         \textbf{TokLIP (Ours)} & LLaVA-v1.5 data  & 256 & 81.2 & 1346.8 & 59.8 & 57.4 & 40.2 ~ \\ 
        \midrule
        Discrete Tokenizer (VQGAN) & LLaVA-v1.5 data  &384 & 69.2 & 773.7 & 34.6 & 41.2 & 34.8  ~ \\ 
        \rowcolor{purple!10} 
         \textbf{TokLIP (Ours)}  & LLaVA-v1.5 data  &384 & 82.7 & 1410.2 & 65.2 & 59.3 & 42.1   ~ \\ 
        \bottomrule
    \end{tabular}
    }
    \end{center}
    \label{tab:vq-clip-mllm-ablation}
    \vspace{-0.2cm}
\end{table*}

\subsection{Comparison with Other Tokenizers}
\label{subsec:exp_tokenizer_compare}

\paragraph{Comprehension capacity.}
We compare TokLIP with state-of-the-art tokenizers~\citep{wu2024vila-u,zhao2025qlip} and CLIP models~\citep{cherti2023openclip,radford2021clip,zhai2023siglip} across both image classification and cross-modal retrieval tasks, shown in Table~\ref{tab:understanding-comparison}.
It can be observed that TokLIP outperforms several SOTA CLIP-based encoders. For instance, TokLIP achieves an impressive 76.4\% accuracy on ImageNet with a 256 input resolution, surpassing MetaCLIP~\citep{xu2023metaclip} and EVA02~\citep{fang2024eva02}.
These results demonstrate that TokLIP effectively inherits semantic understanding from the pre-trained teacher and demonstrates competitive performance on downstream tasks.
Additionally, at both 256x256 and 384x384 input resolutions, TokLIP outperforms VILA-U~\citep{wu2024vila-u} and QLIP~\citep{zhao2025qlip}, particularly excelling in the retrieval task.
TokLIP achieves 68.00\% TR@1 and 52.87\% IR@1, which are 7.14\% and 9.87\% higher than QLIP,  respectively, at a 384 resolution.
This suggests that TokLIP captures more semantic information compared to methods that discretize high-level image features. Notably, VILA-U is trained on COYO-700M~\citep{kakaobrain2022coyo-700m}, and QLIP uses DataComp-1B~\citep{gadre2023datacomp}, both of which involve significantly more resource-intensive training processes than ours.
\hk{
Additionally, TokLIP-XL outperforms SigLIP-B/16, which was trained on the 10B-scale WEBIL dataset, demonstrating that high-resolution inputs combined with a stronger VQGAN tokenizer can further boost comprehension performance.
}
Future work will focus on enhancing TokLIP's performance through the utilization of higher-quality training data~\citep{fang2023dfn,xu2023metaclip}.

\begin{table*}[!t]
    \caption{Comparison with state-of-the-art MLLMs on multimodal comprehension benchmarks. ``Cmp.'' and ``Gen.'' denote ``comprehension'' and ``generation'', respectively. Models using external pretrained diffusion models are marked with $^\dagger$.
    }
    \vspace{-10pt}
    \begin{center}
    \setlength{\tabcolsep}{2pt}
    \renewcommand{\arraystretch}{1.0}
    \resizebox{1\linewidth}{!}{
    \small
    
    \begin{tabular}{lclccccccc}
    \toprule
    \textbf{Model}                                                     & \textbf{\#Param} & \textbf{Res.} & \textbf{POPE$ \uparrow$} & \textbf{MME-P$ \uparrow$} & \textbf{MMB$ \uparrow$} & \textbf{SEED$ \uparrow$} & \textbf{GQA$ \uparrow$} & \textbf{MMMU$ \uparrow$} & \textbf{AI2D$ \uparrow$} \\ \midrule
    \multicolumn{2}{l}{\color{gray}{\textit{Cmp. Only}}} \\
    \rowcolor{gray!10}
    \rowcolor{gray!10}
    InstructBLIP~\cite{instructblip}                                   & Vicuna-7B             & 224           & -                        & -                     & 36.0                  & 53.4                   & 49.2                  & 30.6                        & 33.8                     \\
    \rowcolor{gray!10}
    InstructBLIP~\cite{instructblip}                                   & Vicuna-13B             & 224           & 78.9                   & 1212.8                  & -                     & -                      & 49.5                  & -                       & -                     \\     
    \rowcolor{gray!10}
    Qwen-VL-Chat~\cite{bai2023qwenvl}                                  & Qwen-7B               & 448           & -                        & 1487.5                & 60.6                  & 58.2                   & 57.5                  & 35.9                       & 45.9                          \\
    \rowcolor{gray!10}
    LLaVA-v$1.5$~\cite{liu2024llava1.5}                                & Vicuna1.5-7B           & 336           & 85.9                   & 1510.7                  & 64.3                  & 58.6                   & 62.0                  & 35.4                     & 54.8                    \\    
    \midrule
    
    \multicolumn{2}{l}{\color{gray}{\textit{Cmp. and Gen. Continuous}}} \\
    
    \rowcolor{gray!10}
    DreamLLM$^\dagger$~\cite{dong2023dreamllm}                         & Vicuna1.5-7B           & 224           & -                        & -                     & 58.2                   & -                      & -                     & -                        & -                     \\
    \rowcolor{gray!10}
    LaVIT$^\dagger$~\cite{jin2023unified}                              & LLaMA-7B               & 224           & -                        & -                     & 58.0                  & -                      & 46.8                  & -                        & -                          \\
    \rowcolor{gray!10}
    MetaMorph$^\dagger$~\cite{tong2024metamorph}                       & LLaMA3-8B              & 384           & -                        & -                     & 75.2                  & 71.8                   & -                     & -                        & -                          \\
    \rowcolor{gray!10}
    NExT-GPT$^\dagger$~\cite{wu2023next-gpt}                           & Vicuna-7B              & 224           & -                        & -                     & -                     & 57.5                   & -                     & -                        & -                          \\
    \rowcolor{gray!10}
    SEED-X$^\dagger$~\cite{ge2024seed-x}                               & LLaMA2-chat-13B        & 448           & 84.1                   & 1457.0                  & 70.1                  & 66.5                   & 49.1                  & 35.6                     & -                   \\
    \rowcolor{gray!10}
    ILLUME$^\dagger$~\cite{wang2024illume}                                       & Vicuna1.5-7B           & 224           & 88.5                   & 1445.3                  & 65.1                    & 72.9                   & -                     & 38.2                   & -                     \\
    \rowcolor{gray!10}
    Janus~\cite{wu2024janus}                                           & DeepSeek-LLM-1.5B      & 384           & 87.0                   & 1338.0                  & 69.4                    & 63.7                   & 59.1                  & 30.5                   & -                    \\
    \rowcolor{gray!10}
    Janus-Pro-1B~\cite{chen2025janus-pro}                              & DeepSeek-LLM-1.5B      & 384           & 86.2                   & 1444.0                  & 75.5                    & 68.3                   & 59.3                  & 36.3                   & -                     \\
    \rowcolor{gray!10}
    MAR~\cite{wu2025mar}                                               & Qwen-2.5-1.5B            & 512           & 87.6                  & 1155.0                  & 65.5                    &  67.1                   & 58.9                  & 38.9                   & -                    \\
    %
    \midrule

    \multicolumn{2}{l}{\color{gray}{\textit{Cmp. and Gen. Discrete}}} \\
    Show-o~\cite{xie2024show-o}                                        & Phi-1.5-1.3B           & 256           & 73.8                   & 948.4                   & -                       & -                        & 48.7                  & 25.1                   & -                         \\
    Show-o~\cite{xie2024show-o}                                        & Phi-1.5-1.3B           & 512           & 80.0                   & 1097.2                  & -                       & -                        & 58.0                  & 26.7                   & -                          \\
    LWM~\cite{liu2024lwm}                                              & LLaMA2-7B              & 256           & 75.2                   & -                       & -                       & -                        & 44.8                  & -                      & -                      \\
    D-Dit~\cite{li2024dual}                                            & $2$B diffusion         & 512           & 84.0                   & 1124.7                  & -                       & -                        & 59.2                  & -                      & -                          \\
    Emu$3$-Chat~\cite{wang2024emu3}                                    & $8$B from scratch      & 512           & 85.2                   & 1244.0                  & 58.5                    & 68.2                     & 60.3                  & 31.6                   & 70.0              \\    
    Chameleon~\cite{team2024chameleon}                                 & $7$B from scratch      & 256           & -                      & -                       & -                       & -                        & -                     & 22.4                   & 46.0                      \\
    Orthus~\cite{kou2024orthus}                                        & Chameleon-7B           & 256           & 79.6                   & 1265.8                  & -                       & -                        & 52.8                  & 28.2                    & -                       \\
    VILA-U~\cite{wu2024vila-u}                                         & LLaMA2-7B              & 256           & 83.9                   & 1336.2                  & -                       & 56.3                     & 48.3                  & -                       & -                    \\
    VILA-U~\cite{wu2024vila-u}                                         & LLaMA2-7B              & 384           & \underline{85.8}       & 1401.8                  & -                       & 59.0                     & 60.8      & -                       & -                     \\
    UniTok~\cite{ma2025unitok}                                         & LLaMA2-7B              & 256           & 83.2                   & 1448.0                  & -                      & 61.1                      & -                      & -                        & -                    \\
    UniToken~\cite{jiao2025unitoken}                                   & Chameleon-7B           & 384           & -                      & -                       & 71.1                    & 69.9                     & -                     & 32.8                   & 68.7                        \\
    SemHiTok~\cite{chen2025semhitok}                                   & Vicuna1.5-7B           & 384           & 84.2                   & 1400.6                  & 60.3                    & 62.4                     & \underline{61.0}      & 35.2                   & -                     \\
    SynerGen-VL~\cite{li2024synergen-vl}                               & InternLM2-2.4B         & 512           & 85.3                   & 1381.0                  & 53.7                    & 62.0                     & 59.7                  & 34.2                    & 60.8                    \\
    MUSE-VL~\cite{xie2024muse}                                         & Qwen2.5-7B             & 256           & -                      & -                       & 72.1                    & 69.1                     & -                     & \underline{39.7}                    & 69.8                        \\
    TokenFlow-L~\cite{qu2024tokenflow}                                 & Vicuna1.5-13B          & 256           & 85.0                   & 1365.4                  & 60.3                    & 62.6                     & 60.3                  & 34.4                   & 56.6                     \\
    TokenFlow-XL~\cite{qu2024tokenflow}                                & Vicuna1.5-13B          & 384           & \textbf{86.8}          & \underline{1545.9}      & 68.9                    & 68.7                     & \textbf{62.7}         & 38.7                   & 66.7                     \\
    \rowcolor{purple!10}
    \textbf{TokLIP-L (Ours)}                                            & Qwen2.5-7B             & 384           & 84.9                  & 1496.6                  & \underline{76.9}         & \underline{71.5}        & 57.0               & \textbf{47.1}             & \underline{76.8}                     \\
    \rowcolor{purple!10}    
    \textbf{TokLIP-XL (Ours)}                                            & Qwen2.5-7B             & 512           & 85.2                  & \textbf{1586.8}         & \textbf{77.4}            & \textbf{72.1}           & 57.3               & \textbf{47.1}         & \textbf{77.7}                     \\
    \bottomrule    
    \end{tabular}
    }
    \end{center}
    \label{tab:mllm-understanding}
\end{table*}

\paragraph{Enhanced vision tokenizer for MLLMs.}

To assess the effectiveness of different tokenizer architectures integrated into MLLMs, we compare TokLIP with VILA-U~\citep{wu2024vila-u} and VQGAN~\citep{sun2024llamagen} on multimodal comprehension tasks. Specifically, we use the LLaVA-v1.5~\citep{liu2024llava1.5} framework to evaluate the understanding performance of VQGAN and TokLIP, with VILA-U results reported from the original paper. We train on the LLaVA-Pretrain-558K dataset for MLP pretraining and LLaVA-v1.5-mix-665K for instruction tuning. The comprehensive evaluations are presented in Table~\ref{tab:vq-clip-mllm-ablation}.
Our results demonstrate that TokLIP consistently outperforms VQGAN across all metrics and on nearly all benchmarks. 
For 256 input resolution, TokLIP achieves a clear performance advantage over both VILA-U and VQGAN. Notably, VILA-U is trained on 7M vision-language pairs, significantly more than the LLaVA-v1.5 dataset. 
Despite these variations in training data size, input resolution changes have a negligible impact on VQ-based tokenizers' performance, pointing to the inherent limitations in scaling their comprehension capacity.
These findings further support our hypothesis that: (1) discrete tokenizers struggle with comprehension tasks due to their lack of semantic depth, and (2) discrete semantic tokenizers face performance degradation due to conflicting training objectives. Overall, our results highlight TokLIP's superior ability to capture semantic-level understanding, making it a more effective solution for multimodal comprehension tasks involving discrete inputs.

\subsection{Multimodal Comprehension}
\label{subsec:exp_understanding}

\paragraph{Main results.}
We present a comprehensive evaluation of TokLIP in comparison to state-of-the-art models in both understanding-only MLLMs and unified understanding-and-generation MLLMs. Illustrated in Table~\ref{tab:mllm-understanding}, TokLIP, as a discrete tokenizer, achieves competitive performance when compared to existing methods that rely on continuous inputs from CLIP vision encoders. 
\hk{For instance, TokLIP-L performs on par with LLaVA-v1.5 across MME and POPE, while outperforming it on the SEED Bench, MMB, MMMU, and AI2D datasets.}
These results demonstrate that TokLIP effectively learns semantic features from discrete visual tokens, significantly enhancing multimodal understanding capabilities. By leveraging the pre-trained CLIP model, TokLIP successfully incorporates high-level semantic information, bridging the gap between visual tokens and CLIP-level semantics.

In the context of unified models with discrete inputs, which represents the most comparable setting, TokLIP-L substantially outperforms Chameleon and Show-o across all evaluated benchmarks. 
Moreover, TokLIP-L surpasses Emu3-chat on most benchmarks, despite utilizing much less training data and operating at a lower input resolution (384 vs. 512). This reinforces the effectiveness of TokLIP in successfully semanticizing VQ codes with CLIP-level semantics, showing that high-quality semantic understanding can be achieved with fewer resources.
When compared to VILA-U and TokenFlow, TokLIP-L also delivers superior performance, further supporting our hypothesis that transitioning from discrete to continuous semantics offers a more effective approach, as discussed in Section~\ref{subsec:discussion}. 
\hk{Moreover, TokLIP-XL further boosts comprehension performance across most benchmarks, achieving state-of-the-art results on AI2D, MME, SEED-Bench, and MMB.}
\hk{
The sub-optimal performance on GQA may be attributed to the data mixup strategy in LLaVA-OneVision, and we expect further improvements with training on more QA pairs.
}
These empirical results highlight TokLIP’s potential as a foundational tokenizer for unified understanding and generation tasks, paving the way for more advanced multimodal models.

\paragraph{Data efficiency.}
Approaches~\citep{wang2024emu3,xie2024show-o} that directly optimize discrete VQ-based tokenizers typically rely on large-scale datasets. For example, SynerGen-VL~\citep{li2024synergen-vl} is trained on 170 million samples with task-specific prompts for visual understanding, which is often unaffordable for most researchers. Similarly, discrete semantic models~\citep{wu2024vila-u,qu2024tokenflow,zhao2025qlip} like VILA-U require large-scale training sets to discretize high-level semantics effectively.
In contrast, we emphasize that TokLIP provides an efficient solution for tokenizer training, achieving strong performance with just 80M image-text pairs and fewer resources for visual understanding. 
\hk{Note that both TokLIP and VILA-U are initialized from a pre-trained SigLIP variant, so we exclude the SigLIP pretraining data and discuss solely on the tokenizer training stage.}
This efficiency is due to our pipeline, which bridges the gap from discrete tokens to continuous representations, enabling effective learning with less data and computational overhead.

\subsection{Visual Generation}
\label{subsec:exp_generation}

\paragraph{Class-conditional image generation.}
\lin{
We follow the training pipeline of LlamaGen~\citep{sun2024llamagen} to evaluate the generation capacity on ImageNet benchmarks. 
Specifically, we replace the text-to-image VQGAN tokenizer with our TokLIP tokenizer to train both GPT-Base and GPT-Large models for different epochs, while keeping the training settings and hyperparameters consistent.
We compare the discrete VQGAN tokenizer with TokLIP in Table~\ref{tab:llamagen}. All experiments are conducted with the same sampling configuration: cfg = 2.5, top-k = 0 (all), top-p = 1.0, and temperature = 1.0.
Our results show that all models using the TokLIP tokenizer achieve a lower FID than those using the discrete VQGAN tokenizer, demonstrating that the incorporation of high-level semantics enhances image generation. 
The improvement is less significant at the 256 input image size compared to the 384 resolution, which we attribute to the fact that semantics are more beneficial when dealing with longer image tokens.
}

\paragraph{Qualitative evalutaion.}

We show class-conditional generation examples with TokLIP in Figure~\ref{fig:gene}. Images are generated at 384x384 and resized to 256x256 for display.
It can be observed that the model generates high-quality, aesthetically pleasing images that effectively capture visual concepts.

\begin{table*}[h]
\begin{minipage}{.42\textwidth}
\caption{Performance on class-conditional 256x256 ImageNet 50K benchmark. The downsample ratio is 16.
}
\label{tab:llamagen}
\begin{center}
\resizebox{\linewidth}{!}{
\begin{tabular}{llccc}
\toprule
\textbf{Tokenizer} & \textbf{Model} & \textbf{Res.} & \textbf{epochs} & \textbf{FID $\downarrow$} \\ \midrule
VQGAN              & GPT-B          & 256          & 50              & 13.21        \\
TokLIP             & GPT-B          & 256          & 50              & 13.12        \\
VQGAN              & GPT-B          & 384          & 50              & 14.48        \\
TokLIP             & GPT-B          & 384          & 50              & 12.37        \\ \midrule
VQGAN              & GPT-B          & 256          & 300             & 7.43         \\
TokLIP             & GPT-B          & 256          & 300             & 7.29         \\
VQGAN              & GPT-B          & 384          & 300             & 9.70         \\
TokLIP             & GPT-B          & 384          & 300             & 7.64         \\ \midrule
VQGAN              & GPT-L          & 256          & 50              & 5.96         \\
TokLIP             & GPT-L          & 256          & 50              & 5.72         \\
VQGAN              & GPT-L          & 384          & 50              & 7.28         \\
TokLIP             & GPT-L          & 384          & 50              & 6.19         \\ \bottomrule
\end{tabular}
}
\end{center}
\end{minipage}
\hspace{10pt}
\begin{minipage}{0.5\textwidth}

\begin{center}
\captionof{figure}{Visualizations of class-conditional generation by TokLIP using LlamaGen-B framework.
}
\label{fig:gene}
\vspace{-5pt}
\includegraphics[scale=0.25]{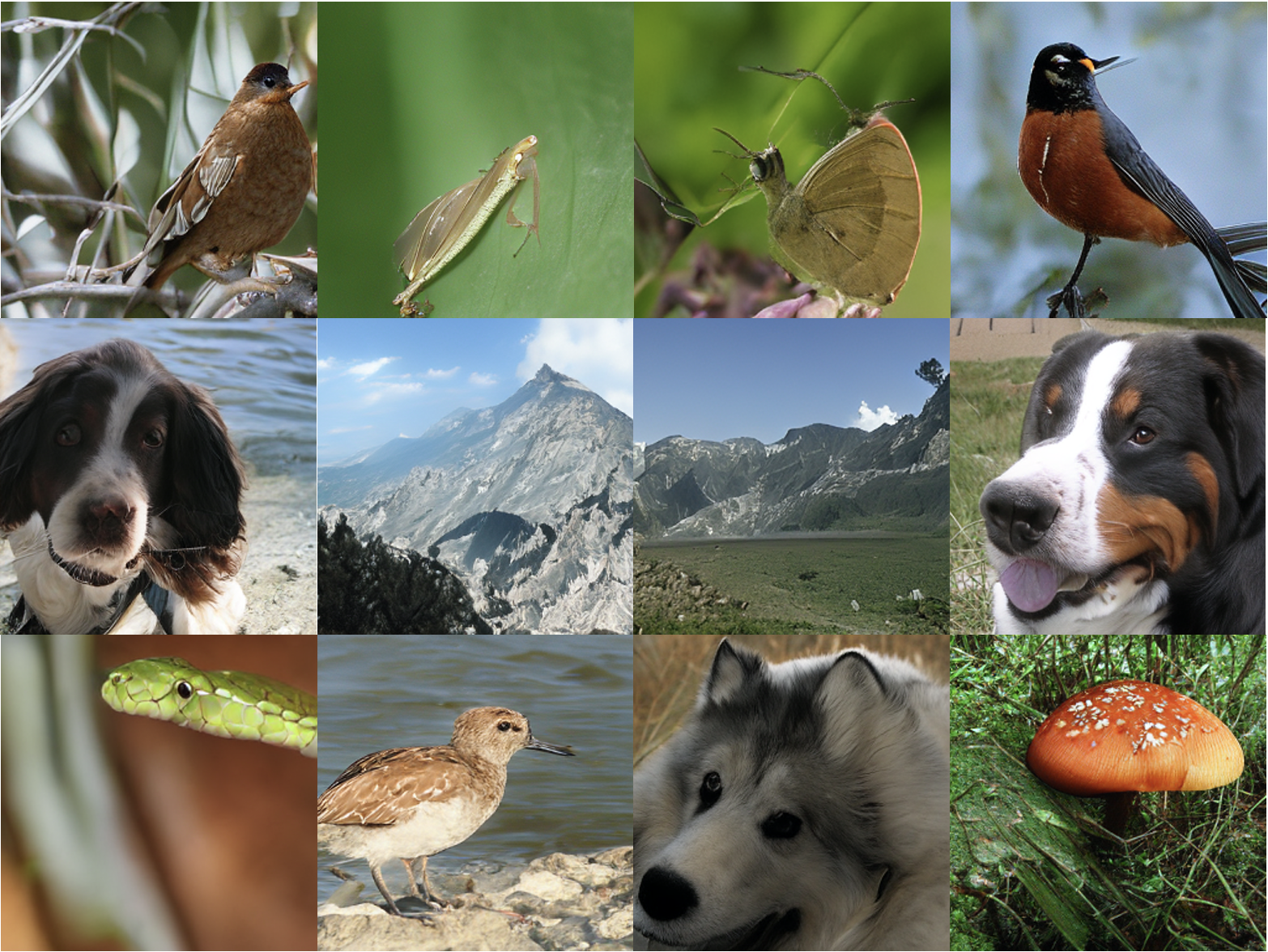}
\end{center}
\end{minipage}
\end{table*}

\subsection{Ablation Studies}
\label{subsec:ablation}

\paragraph{TokLIP training.}
We explore several training strategies for our TokLIP models. Specifically, we evaluate three key aspects: (1) whether to initialize the vision encoder with pre-trained weights, (2) the method for projecting VQ codes into the CLIP feature space, and (3) the design of the learning objectives.
To analyze these strategies, we implement various ablation models based on OpenCLIP ViT-B/16~\citep{cherti2023openclip} and pre-train on CC3M dataset~\citep{changpinyo2021conceptual}, listing
the results in Table~\ref{tab:TokLIP-ablation}.

For \textbf{vision initialization}, omitting the pre-trained vision encoder significantly degrades performance, as large-scale pre-training provides the CLIP encoder with a deep understanding of image semantics, enabling it to capture complex patterns and textures. By leveraging pre-trained weights, TokLIP benefits from this rich knowledge, improving its semantic comprehension capabilities. 

For \textbf{projection methods}, we compare MLP projection with a direct codebook-based mapping of discrete tokens into the CLIP feature space. As shown in Table~\ref{tab:TokLIP-ablation}, MLP projection outperforms codebook mapping, likely due to the large codebook size (16,532 entries), which complicates the learning of meaningful representations. 
A more compact or optimized codebook could enhance performance by focusing on a more relevant set of visual features, a direction we leave for future work.

Regarding \textbf{learning objectives}, our results show that distillation from a teacher model significantly enhances TokLIP's performance. Distilling the [CLS] token features from the teacher improves understanding of image-text relationships, while distilling all tokens does not lead to gains. This suggests that the [CLS] token, representing global input understanding, is most beneficial for distillation. The lack of improvement when distilling all tokens may stem from the differing token dependencies between the causal attention mechanism in TokLIP and the bidirectional dependencies in the teacher model, leading to misalignment in token-level representations.

Based on these empirical findings, we adopt a training strategy that loads pre-trained vision encoder weights to retain semantic knowledge, employs MLP projection for efficient code mapping, and integrates contrastive learning with knowledge distillation for effective learning.

\begin{table}[!t]
    \caption{Ablations of different vision initialization, projection methods and learning objectives for training TokLIP on CC3M.}
    \begin{center}
    \resizebox{0.8\linewidth}{!}{
    \begin{tabular}{ccccc}
    \toprule
    \textbf{Vision Encoder} & \textbf{Mapping}    & \textbf{Training Objective}                         & \textbf{Top 1 Acc.}    & \textbf{Top 5 Acc.}    \\ \midrule
    Init                    & Patch               & $\mathcal{L}_\text{contra}$                              & 18.50                  & 39.97                \\
    Init                    & MLP              & $\mathcal{L}_\text{contra}$                              & 15.73                  & 36.81                \\
    ViT-B/16                 & MLP              & $\mathcal{L}_\text{contra}$                              & 32.93                  & 60.33                \\
    ViT-B/16                 & Soft MLP         & $\mathcal{L}_\text{contra}+\mathcal{L}_\text{distill}$         & 51.53                  & 79.85                \\
    ViT-B/16                 & Codebook            & $\mathcal{L}_\text{contra}+\mathcal{L}_\text{distill}$         & 35.17                  & 65.07                \\
    ViT-B/16                 & MLP              & $\mathcal{L}_\text{contra}+\mathcal{L}_\text{distill\_all}$  & 51.04                  & 79.29                \\
    ViT-B/16                 & MLP              & $\mathcal{L}_\text{contra}+\mathcal{L}_\text{distill}$         & 52.46                  & 80.52                \\ \bottomrule
    \end{tabular}
    }
    \end{center}
    \label{tab:TokLIP-ablation}
\end{table}

\paragraph{Feature fusion.}
\lin{
We explore different fusion functions to complement the low-level tokens from VQGAN with semantic-level features from the causal token encoder. Specifically, we compare the following strategies: (1) direct sum, (2) weighted sum with a learnable parameter, and (3) concatenation within the LlamaGen-Base (384x384) framework. We evaluate all models after training for 50 epochs, generating 10,000 samples for quick evaluation. As shown in Table~\ref{tab:fusion}, concatenation yields the lowest FID, which we adopt as the final approach. Additionally, all fusion functions outperform the discrete VQGAN, further demonstrating that incorporating high-level semantics enhances visual generation capabilities.
}

\begin{table}[!t]
    \caption{Ablations of fusion function for pixel-level features and semantic-level features using LlamaGen-B.
    The inference setting is top-k = 0 (all), top-p = 1.0, temperature = 1.0 for all experiments.
    }
    \begin{center}
    \resizebox{0.85\linewidth}{!}{
    \begin{tabular}{lccccc}
    \toprule
    \textbf{Method}       & \textbf{Input\_size} & \textbf{Epochs} & \textbf{Cfg\_ratio} & \textbf{Generated\_samples} & \textbf{FID $\downarrow$} \\ \midrule
    VQGAN                 & 384                 & 50             & 2.5                 & 10K                        & 17.31        \\ 
    TokLIP\_sum           & 384                 & 50             & 2.5                 & 10K                        & 16.37        \\
    TokLIP\_weighted\_sum & 384                 & 50             & 2.5                 & 10K                        & 15.46        \\
    TokLIP\_concatenation & 384                 & 50             & 2.5                 & 10K                        & 14.93        \\ \bottomrule
    \end{tabular}
    }
    \end{center}
    \label{tab:fusion}
\end{table}



\subsection{Fine-grained Evaluation}
\label{subsec:fine-grained}

\hk{
To further assess the generalization capacity of the TokLIP tokenizer, we conduct detailed evaluations on fine-grained zero-shot classification tasks. We compare TokLIP with QLIP~\citep{zhao2025qlip} across nine widely used benchmarks, including Caltech101~\citep{fei2004Caltech101}, Flowers102~\citep{nilsback2008flowers}, Oxford Pets~\citep{parkhi2012cats}, Stanford Cars~\citep{krause2013cars}, FGVC Aircraft~\citep{maji2013aircraft}, SUN397~\citep{xiao2010sun397}, and Food101~\citep{bossard2014food}.
As shown in Table~\ref{tab:reb_class}, TokLIP consistently outperforms QLIP on all datasets, achieving a higher average accuracy (85.03 vs. 78.88), highlighting its strong generalization. These results further validate the effectiveness of our semanticization pipeline for VQ tokens.
}

\begin{table*}[!h]
\caption{TokLIP performance on fine-grained classification benchmarks.}
\label{tab:reb_class} 
\centering
\resizebox{0.85\linewidth}{!}{
\begin{tabular}{cccccccccc}
\toprule
\textbf{Model}  & \textbf{Res.} & \textbf{Pets}  & \textbf{Flowers} & \textbf{Food101} & \textbf{SUN397} & \textbf{Cars}  & \textbf{CalTech101} & \textbf{Aircraft}  & \textbf{Avg.} \\ \midrule
QLIP   & 392 & 91.03 & 78.63   & 89.66   & 72.29  & 88.25 & 97.32      & 34.95    & 78.88       \\
\rowcolor{purple!10}
TokLIP & 384 & 93.64 & 88.58   & 89.98   & 74.77  & 93.31 & 97.65      & 57.25   & 85.03         \\ \bottomrule
\end{tabular}
}
\end{table*}

\hk{
Considering fine-grained capacity in MLLMs, TokLIP demonstrates a strong ability to understand scientific diagrams, as evidenced by its superior performance on the AI2D dataset. 
However, VQ tokens inherently suffer from information loss due to quantization, which is a well-known limitation for discrete MLLMs~\citep{xie2024show-o,team2024chameleon,wang2024emu3}. 
We freeze the VQ encoder to ensure extensibility to stronger VQ tokenizers, but this design may cause a certain performance drop when dealing with text-rich figures, where fine-grained text information is small. 
This limitation mainly originates from the VQ tokenizer itself~\citep{shi2024ibq,agarwal2025cosmos}. 
Several works~\citep{li2024synergen-vl,xie2025show-o2,wang2024emu3,wang2025ett} have explored ways to mitigate this issue, such as training with high-resolution images, utilizing text-rich datasets, or unlocking the VQ tokenizers. 
Reinforcement learning has also proven to be an effective method for enhancing the capabilities of unified MLLMs in both fine-grained generation and understanding tasks~\citep{geng2025xomni,xiao2025mindomni,jiang2025t2i}.
TokLIP can be further enhanced by adopting these strategies, and we plan to explore them in the future work.
}

\subsection{Bidirectional vs. Causal Attention}
As introduced in Section~\ref{subsec:toklip_method}, we apply a causal attention module in TokLIP for unified understanding and generation. 
Our causal attention-based TokLIP demonstrates superior semantic understanding compared to the VILA-U. 
In this section, we present a detailed analysis of bidirectional and causal attention mechanisms, examining their respective impacts on comprehension and generation tasks.

\paragraph{Bidirectional attention enhances comprehension.}
We first implement a bidirectional attention version using OpenCLIP-ViT-B/16~\citep{cherti2023openclip} to investigate whether a bidirectional architecture could further enhance comprehension tasks.
We pre-train both the causal and bidirectional models on the CC3M and CC12M datasets with the same training setups.
Results in Table~\ref{tab:bidirectional} show that bidirectional-based models surpass causal-based models on classification and retrieval tasks. 
This suggests that a bidirectional TokLIP can further improve comprehension capacity for MLLMs, reinforcing the idea that learning semantics from discrete tokens is more effective than discretizing continuous features.

\begin{table}[!h]
    \caption{Ablations on bidirectional and causal attention in TokLIP.}
    \begin{center}
    \resizebox{0.8\linewidth}{!}{
    \begin{tabular}{cccccc}
    \toprule
    \textbf{Vision Encoder} & \textbf{Attention} & \textbf{Dataset} & \textbf{IN Top1$ \uparrow$} & \textbf{COCO TR@1$ \uparrow$} & \textbf{COCO IR@1$ \uparrow$} \\ \midrule
    ViT-B/16                 & Bidirectional             & CC3M                   & 56.32                  & 44.66                & 32.66                \\
    ViT-B/16                 & Causal                    & CC3M                   & 52.46                  & 41.30                & 30.13                \\ \midrule
    ViT-B/16                 & Bidirectional             & CC12M                  & 58.99                  & 47.32                & 33.78                \\
    ViT-B/16                 & Causal                    & CC12M                  & 55.02                  & 42.80                & 31.06                \\ \bottomrule
    \end{tabular}
    }
    \end{center}
    \label{tab:bidirectional}
    \vspace{-0.2cm}
\end{table}

\begin{wrapfigure}{r}{7cm}
    \caption{
    Illustration of non-causal info for bidirectional TokLIP and quantitative generation results.
    }
    \includegraphics[width=1\linewidth]{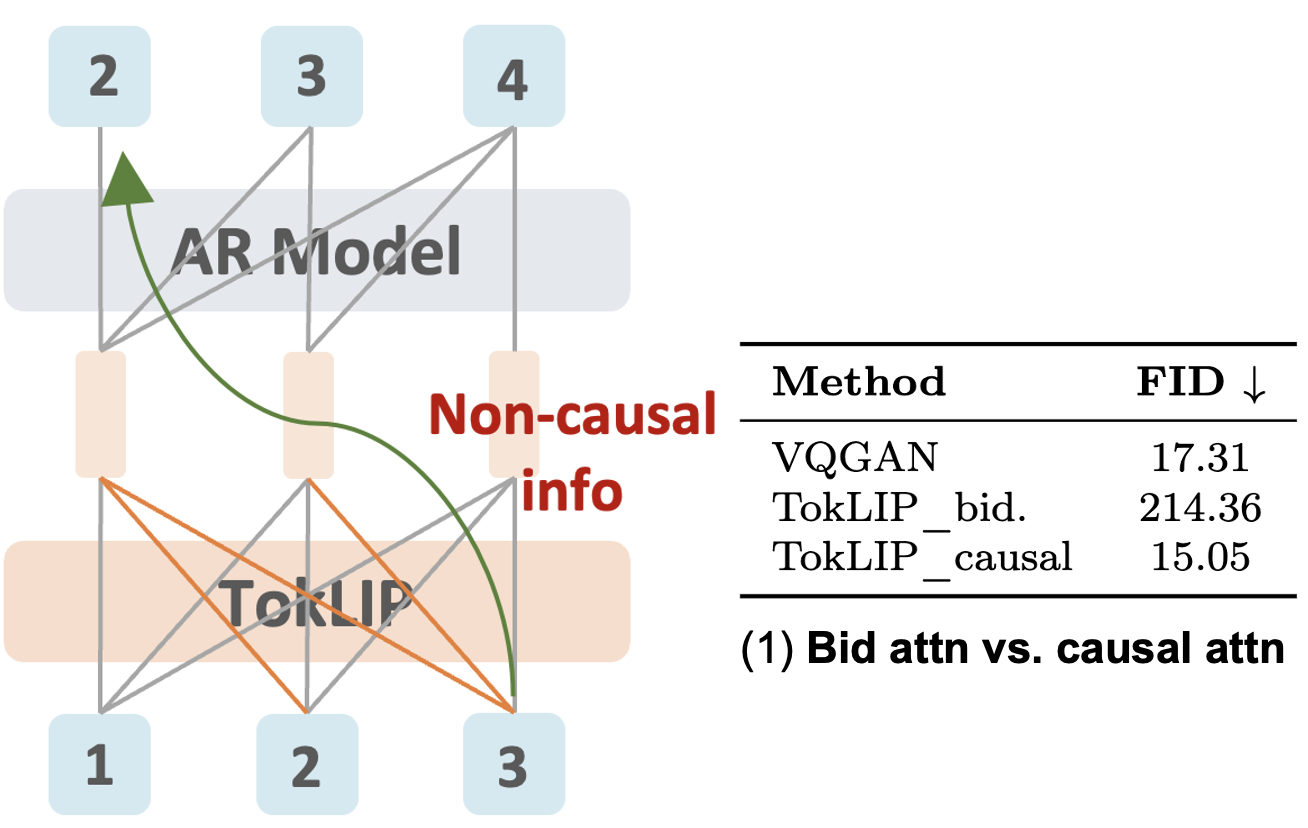}
    \label{fig:pruning_impact}
    \vspace{-15pt}
\end{wrapfigure}

\paragraph{Causal attention underpins generation.}
Causal TokLIP is necessary for generation tasks: 
1) TokLIP with bidirectional attention introduces non-causal information flow as shown in Figure~\ref{fig:pruning_impact}, which breaks the autoregressive nature for next-token prediction. 
2) 
We conduct ablation of causal vs. bidirectional TokLIP. Both models are trained on CC3M using the LlamaGen-B framework.
Bidirectional TokLIP (FID 214.36 on ImageNet) performs significantly worse than causal one (15.05) and baseline VQGAN (17.31). Performance degradation lies in the gap between non-causal training and autoregressive inference, which confirms that causal attention is essential.

Based on the detailed analysis and experimental results on both comprehension and generation tasks, we conclude that causal attention is the preferred choice for TokLIP. This decision is motivated by its ability to preserve autoregressive generation capability, which is essential for maintaining a unified architecture.
\section{Conclusion}
\label{sec:conclusion}

In this paper, we introduce TokLIP, a discrete-to-continuous visual tokenizer that enables end-to-end autoregressive training of a single Transformer with sequences of multimodal discrete tokens. Our approach offers three key advantages: it disentangles the training objectives for comprehension and generation, enhances comprehension performance, and efficiently uses state-of-the-art VQ tokenizers. Through empirical validation, we demonstrate that TokLIP achieves exceptional image representation capabilities, reduces training cost, and \lin{enhances} image generation capability. Our results underscore the potential of TokLIP to integrate diverse abstraction levels for robust multi-modal outputs.

\newpage

\clearpage

\appendix

\renewcommand\thefigure{\Alph{section}\arabic{figure}}
\renewcommand\thetable{\Alph{section}\arabic{table}}
\setcounter{figure}{0}
\setcounter{table}{0}

\section*{Appendix}

\section{Additional Implementation Details}

\paragraph{Hyperparameters for TokLIP.}

We list detailed hyperparameters for training TokLIP in Table~\ref{tab:hyper_toklip}.

\begin{table}[!h]
\caption{Detailed Hyperparameters for training TokLIP.}
\begin{center}
\resizebox{0.55\linewidth}{!}{
\begin{tabular}{l|cc}
\toprule
Config              & TokLIP    &   TokLIP          \\ 
\midrule
Resolution          & 256x256            &  384x384                     \\
Optimizer          & \multicolumn{2}{c}{AdamW}                                     \\
Optimizer momentum & \multicolumn{2}{c}{$\beta_1$=0.9, $\beta_2$=0.98}             \\
LR schedule        & \multicolumn{2}{c}{CosineLRScheduler}                         \\
Weight decay       & \multicolumn{2}{c}{0.1}                                      \\
Warmup steps       & \multicolumn{2}{c}{500}                                       \\
Base LR            & 1e-5            & 1e-5                                        \\
Batch size         & 1792            & 512                                        \\
\bottomrule
\end{tabular}
}
\end{center}
\label{tab:hyper_toklip} 
\vspace{-0.5cm}
\end{table}

\paragraph{Mapping function.}
For the multi-layer perceptron projection between VQGAN and token encoder, we first map the 8-dimensional features from VQGAN to 4×dimensional hidden states of CLIP features, followed by a GeLU activation layer, and then use another linear layer to map to the CLIP feature dimension.
For the codebook approach, we directly establish a large embedding layer to map each code to the CLIP feature space.

\section{More Experimental Results}

\subsection{Further Discussion}

\paragraph{Effects of all token distillation on bidirectional TokLIP.}
As analyzed in Section~\ref{subsec:ablation}, distilling features from all tokens does not enhance causal-attention TokLIP due to different attention dependencies. Here, we ablate the all-token distillation loss for training bidirectional-attention TokLIP on CC3M dataset. As shown in Table~\ref{tab:appx_bi_alltoken}, distilling all tokens significantly improves performance on downstream tasks.
This indicates that bidirectional-attention TokLIP shares a more similar architecture with the teacher model and thus inherits more semantic knowledge. This provides a promising approach to better empower discrete tokens with semantics for multimodal understanding tasks.

\begin{table}[!h]
    \caption{Ablation of all token distillation for bidirectional TokLIP.}
    \begin{center}
    \resizebox{0.9\linewidth}{!}{
    \begin{tabular}{cccccc}
    \toprule
    \textbf{Vision Encoder} & \textbf{Attention} & \textbf{Distillation} & \textbf{IN Top1$ \uparrow$} & \textbf{COCO TR@1$ \uparrow$} & \textbf{COCO IR@1$ \uparrow$} \\ \midrule
    ViT-B/16                 & Bidirectional      & Single token          & 56.32            & 44.66              & 32.66              \\
    ViT-B/16                 & Bidirectional      & All token             & 59.47            & 47.32              & 35.60              \\ \bottomrule
    \end{tabular}
    }
    \end{center}
    \label{tab:appx_bi_alltoken}
    \vspace{-0.5cm}
\end{table}

\subsection{Effects of TokLIP with small LLMs.}
\hk{
To further evaluate the effectiveness of TokLIP, we train TokLIP-XL with a smaller LLM, Qwen-2.5-1.5B, using the LLaVA-v1.5 dataset.
We compare our model with MAR~\citep{wu2025mar}, which uses the same LLM but is trained on a significantly larger dataset (29.8M vs. 1.2M).
As shown in Table~\ref{tab:appx_harmon}, TokLIP achieves competitive results compared to MAR, even with much fewer data sources.
This showcases the effectiveness of TokLIP even when paired with smaller-scale language models.
}

\begin{table}[h]
\caption{Comprehension results with Qwen2.5-1.5B-Instruct.}
\begin{center}
\resizebox{0.75\linewidth}{!}{
\begin{tabular}{llcccccc}
\toprule
Model        & Data                 & POPE & MME-P  & MMB  & SEED & MMMU  \\ \midrule
MAR-H        & 25M pt+Onevison sft  & 87.6 & 1155.0 & 65.5 & 67.1 & 38.9  \\
\rowcolor{purple!10}
TokLIP       & LLaVA-v1.5           & 84.4 & 1222.0 & 64.5  & 64.7  & 39.4  \\ 
\bottomrule
\end{tabular}
}
\end{center}
\label{tab:appx_harmon} 
\vspace{-0.5cm}
\end{table}

\newpage

\bibliography{nips}
\bibliographystyle{plainnat}

\end{document}